\documentclass[preprint]{article}

\usepackage[dvipsnames]{xcolor}

    \PassOptionsToPackage{numbers, compress}{natbib}

\usepackage{neurips_2024}

\usepackage{microtype}
\usepackage{booktabs} 

\usepackage{fancyvrb}
\usepackage[utf8]{inputenc}
\usepackage[T1]{fontenc}
\usepackage{url}
\usepackage{graphicx,wrapfig,lipsum}
\usepackage{amsfonts}
\usepackage{nicefrac}
\usepackage{microtype}
\usepackage{dsfont}
\usepackage{subcaption}
\usepackage{MnSymbol}
\usepackage{algorithm}
\usepackage[noend]{algorithmic}
\usepackage{booktabs,tabularx, multirow}
\usepackage{tikz}
\usetikzlibrary{bayesnet, positioning, shapes, arrows}
\usepackage{mathtools}
\usepackage{amsmath,amsthm}
\usepackage{soul}
\usepackage{blindtext}
\usepackage{tcolorbox}
\usepackage{todonotes}
\usepackage{thmtools}
\usepackage{wrapfig}

\definecolor{mydarkblue}{rgb}{0,0.08,0.45}
\usepackage[utf8]{inputenc} 
\usepackage[T1]{fontenc}    
\usepackage[colorlinks=true,
    linkcolor=mydarkblue,
    citecolor=mydarkblue,
    filecolor=mydarkblue,
    urlcolor=mydarkblue]{hyperref}
\usepackage{url}            
\usepackage{booktabs}       
\usepackage{nicefrac}       
\usepackage{microtype}      
\usepackage{xcolor}         
\usepackage{enumitem}
\usepackage{amsmath}
\usepackage{graphicx}
\usepackage{tikz}
\usetikzlibrary{bayesnet}
\usepackage{amsfonts}       

\usepackage{xr}

\newtheorem{defn}{Definition}[section]

\title{Simulation-based Benchmarking for Causal Structure Learning in Gene Perturbation Experiments}

\author{%
  Luka Kova\v{c}evi\'{c} \\
  MRC Biostatistics Unit\\
  University of Cambridge\\
  Cambridge, UK \\
  \texttt{luka.kovacevic@mrc-bsu.cam.ac.uk} \\
  \And
  Izzy Newsham\\
  MRC Biostatistics Unit\\
  University of Cambridge\\
  Cambridge, UK \\
  \AND
  Sach Mukherjee \\
  DZNE and University of Bonn, Bonn, Germany,\\
  MRC Biostatistics Unit\\
  University of Cambridge,
  Cambridge, UK \\
  \And
  John Whittaker \\
  MRC Biostatistics Unit\\
  University of Cambridge\\
  Cambridge, UK \\
}

\newcommand{\ltkiz}{25pt}

\begin{document}

\maketitle

\begin{abstract}
    Causal structure learning (CSL) refers to the task of learning causal relationships from data. Advances in CSL now allow learning of causal graphs in diverse application domains, which has the potential to facilitate data-driven causal decision-making. Real-world CSL performance depends on a number of \textit{context-specific} factors, including context-specific data distributions and non-linear dependencies, that are important in practical use-cases. However, our understanding of how to assess and select CSL methods in specific contexts remains limited. To address this gap, we present \textit{CausalRegNet}, a multiplicative effect structural causal model that allows for generating observational and interventional data incorporating context-specific properties, with a focus on the setting of gene perturbation experiments. Using real-world gene perturbation data, we show that \textit{CausalRegNet} generates accurate distributions and scales far better than current simulation frameworks. We illustrate the use of CausalRegNet in assessing CSL methods in the context of interventional experiments in biology.
\end{abstract}

\section{Introduction}

The aim of causal structure learning (CSL) is to infer causal relationships between variables, typically in the form of a directed graph. CSL has been extensively studied and a wide array of methods are available in the literature \citep{heinze2018}.
Recent years have seen considerable 
theoretical and methodological 
advances in CSL, including 
reformulation via continuous optimization \citep{zheng2018dags}, 
neural methods \citep{lippe2021,lopez2022large}
and emerging, end-to-end supervised deep learning-based CSL approaches \citep{ke2022,lagemann2023}.

However, despite these exciting developments, CSL 
in scientific applications remains challenging due to a number of factors, including problem dimension, noise, data limitations and latent variables. Although many CSL methods are supported by theory, in practice it is challenging to verify underlying assumptions. Thus, it is often difficult to assess the likely efficacy of a CSL approach in a  real problem context. This means that in many contexts --  often involving arduous and costly experiments -- it remains
nontrivial to judge whether a given CSL approach is well suited to the setting.


The application of CSL to problems in biomedicine manifests many such problems. Advances in cell and molecular biology have dramatically improved our ability to perform interventional experiments at scale. Notably, the combination of gene editing technologies (e.g. CRISPR) with sequencing protocols \citep{dixit2016perturb} has enabled measurement of tens of thousands of variables under thousands of interventions, with recent experiments spanning millions of cells \citep{replogle2022mapping}. These developments are in the process of transforming biological and biomedical discovery, enabling causal experiments at an unprecedented scale. Large-scale gene knockout experiments in particular, promise to inform drug target identification by allowing for inference of the effects of individual and combinatorial genetic interventions. An emerging body of work has focused on CSL for biological data
\citep[among others][]{hill2016, lopez2022large, lagemann2023},
but for a given problem setting it remains practically challenging to understand 
efficacy and therefore to establish reliable CSL-based workflows.

\paragraph{Related Work.} 
The need to better understand CSL performance characteristics 
has inspired a line of work in benchmarking methods \citep{eigenmann2020, chevalley2022causalbench}. 
While a number of papers include evaluation using real biological data
\citep{hill2016, lopez2022large, lagemann2023},
these results are inevitably specific to the particular real datasets used and relative performance of CSL methods can vary significantly depending on context.
This has motivated work on simulation engines of various kinds. 
In the biological domain, Single-cell Expression of Genes In-silico (SERGIO) \cite{dibaeinia2020sergio}  was developed as an approach based on stochastic differential equations.
More recently, GRouNdGAN \cite{zinati2024groundgan} was proposed as a technique for generating single-cell RNA sequencing (scRNA-seq) data from a given causal graph. CausalBench \citep{chevalley2022causalbench} has also been proposed as a framework for evaluating structure learners directly from real-world data using gene regulatory networks derived from literature and statistical hypothesis testing. 

Large-scale data simulation is an important part of the deep learning CSL method reported in 
\cite{ke2022}. Nonetheless, our approach could be used to generate data to train, or pre-train, supervised CSL methods, a point we return to in the \hyperref[sec:discussion]{Discussion}.

\paragraph{Contributions.} We propose CausalRegNet, an approach for generating realistic synthetic data 
from an underlying directed acyclic graph (DAG).
CausalRegNet allows generation of both observational and interventional data whilst respecting user-specified statistical properties and conforming to a number of basic desiderata. Crucially, CausalRegNet is highly scalable, allowing study of problems of the  size now seen in contemporary experiments and data acquisition pipelines. This simulator can be used to study the performance of CSL methods in a context-specific manner, by fixing key aspects of the problem setting. 

Our work has several novel contributions:
\begin{enumerate}[label=(\roman*)]
    \item We present a novel tool called \textit{CausalRegNet}\footnote{Code is available at  \href{https://github.com/luka-kovacevic/causalregnet}{\Verb+https://github.com/luka-kovacevic/causalregnet+}.} for simulating realistic scRNA-seq data with respect to a causal graph;
    \item We validate CausalRegnet using data from real world large-scale interventional experiments in biology;
    \item We demonstrate how CausalRegNet can be used to benchmark CSL algorithms in practice.
\end{enumerate}

The methods, software and workflow we provide allow for assessment of CSL methods in the context of real-world problems. Our aim is \textit{not} to provide a comprehensive empirical benchmarking study but rather demonstrate how such a benchmark can be carefully devised whilst taking into account many context-specific factors and criteria. Our simulator generates data that behaves similarly to data from (specific) real experiments, allowing for inferences that could not be made using an entirely generic simulation model.

\section{Background} \label{sec:background}

In Section \ref{sec:scm} we introduce the general background knowledge needed for structural causal models (SCMs). This is followed by the introduction of additive noise models (ANMs) in Section \ref{sec:anm}, which are a specific subclass of SCMs that are commonly used to generate data for comparative benchmarks of CSL methods.

\subsection{Structural Causal Models}  \label{sec:scm}

\begin{defn}[Structural Causal Model; \cite{peters2017elements}] A \textit{structural causal model} (SCM) \(\mathfrak{C} = (S, \mathcal{P}_N)\) is a collection \(S\) of \(d\) structural assignments corresponding to each \(X_j \in \mathbf{X} = \{X_1, \ldots, X_d\}\), \begin{equation}
    X_j := f_j(X_{\text{pa}(j)}, N_j), \;\;\;\;\;\; j=1,\ldots,d,
\end{equation} where \(X_{\text{pa}(j)} \subseteq \mathbf{X}\backslash \{X_j\} \) are \textit{parents} of $X_j$, \(\mathcal{P}_N\) is a joint distribution over noise variables, which we require to be jointly independent, such that we can write \(\mathcal{P}_N\) as a product of the marginal densities of \(N_j\) for \(j=1,\ldots,N\). If \(X_i \in X_{\text{pa}(j)}\) then \(X_j\) is a \textit{child} of \(X_i\) and \(X_i\) is a \textit{parent} of \(X_j\).
\end{defn} 

The causal graph \(\mathcal{G}=(V,E)\) is composed of vertices \(V=[d]\) and edges \(E=\{i \rightarrow j : X_i \in X_{\text{pa}(j)}\}\). We assume that \(\mathcal{G}\) has no paths of the form \(j \rightarrow \ldots \rightarrow i\), that is, \(\mathcal{G}\) is acyclic and there are no paths along directed edges that lead from a child to one of its parents. For any node \(j\) if there exists a (\textit{directed}) path \(i \rightarrow \ldots \rightarrow j\) along directed edges then node \(i\) is an ancestor of \(j\). Conversely, node \(j\) is a descendant of node \(i\).

Thus, the observational distribution over $\mathbf{X}$ implied by an SCM is such that each $X_j=f_j(X_{\text{pa}(j)}, N_j)$ in distribution. Assuming there are no interventions, the distribution entailed by the SCM is the \textit{observational} distribution, also more generally called an \textit{entailed} distribution. 

\paragraph{Interventions.} In the SCM framework, interventions can be performed on individual nodes or groups of nodes, where the set of intervention targets is denoted by $I \subseteq V$. When considering more than one intervention $I$, the interventional family $\mathcal{I}:=\{I_1,\ldots, I_K\}$ contains $K$ distinct intervention settings. The observational setting can be denoted as $I^\emptyset$. The distribution entailed by the SCM following an intervention is referred to as an \textit{interventional} distribution.

Interventions are difficult to carry out in practice and at best limited in scope in many settings. 

However, in many molecular biology settings interventional experiments are possible, e.g. by interventions targeted at specific molecular mechanisms. Ongoing technological developments, including methods rooted in  CRISPR and related gene editing protocols, are now enabling large-scale interventional experiments including those further described in Appendix \ref{sec:scrnaseq}.

\subsection{Additive Noise Models} \label{sec:anm}

\begin{defn}[Additive Noise Model; \cite{hoyer2008nonlinear}]

An SCM with structural assignments of the form, \begin{equation}
    X_j := f_j(X_{\text{pa}(j)}) + N_j
\end{equation} where \(f_j(\cdot)\) is an arbitrary function and the noise variables \(N_j\) are jointly independent, is an additive noise model (ANM).
\end{defn}

The case where the function \(f_j(\cdot)\) is linear and the noise variables are distributed as \(N_j \sim \mathcal{N}(0, \sigma^2_j)\) yields the general family of linear Gaussian ANMs. We refer to these simply as Gaussian ANMs throughout.

Synthetic data generated through linear ANMs has previously been shown to be vulnerable to CSL methods that exploit \textit{varsortability} to improve causal structure learning performance.

\begin{defn}[Varsortability; \cite{reisach2021beware}]
    Varsortability, denoted by $v$, is a measure of the agreement between the order of increasing marginal variance and the causal order, \begin{equation}
    v := \frac{\sum_{k=1}^{d-1} \sum_{i \rightarrow j \in E^k} \text{\Verb+increasing+}(\text{Var}(X_i), \text{Var}(X_j))}{\sum_{k=1}^{d-1} \sum_{i \rightarrow j \in E^k} 1},  
\end{equation} \begin{equation}
    \text{where \Verb+increasing+} (a, b) = \begin{cases}
        1 & a < b, \\
        1/2 & a = b, \\
        0 & a > b.
    \end{cases}
\end{equation}
\end{defn}

Methods that score close to 1 or close to 0 are varsortable, meaning the correct causal ordering can be obtained by ordering the nodes by ascending or descending variance, respectively. 

\section{Methodology} \label{sec:method}

To translate recent developments in CSL to the real-world inference of causal structures, it is essential that we understand exactly when CSL algorithms fail and why. This type of understanding requires both benchmarking studies and theoretical work that accounts for context-specific data properties. Motivated by the foregoing, we focus on context-specific simulation for CSL. In Section \ref{sec:desiderata}, we begin by deriving simulator desiderata that are important to a context-specific simulation tool for large-scale gene perturbation experiments. Section \ref{sec:framework} describes the model developed based on desiderata, including an example of conditions relating to the functional relationship between parents and children that ensures our model adheres to domain knowledge of observational and interventional relationships present in scRNA-seq data.

\subsection{Desiderata} \label{sec:desiderata}

The desiderata applied to develop CausalRegNet are as follows: \begin{enumerate}[label=\Roman*.]
    \item \textbf{Interpretability}\label{des1}: Simulators that are parametrically interpretable are easier to use and thus allow for simpler reasoning when it comes to comparing models.
    \item \textbf{Scalability}\label{des2}: Contemporary datasets can be very high dimensional, motivating a need for scalability. Scalability is particularly important for simulation datasets that are generated from a large set of graph configurations as needed for model validation or supervised model training. 
    \item \textbf{Low varsortability}\label{des3}: Data generated by simple ANMs is highly varsortable. Previous work by \cite{reisach2021beware} has established that several CSL algorithms are able to exploit varsortability to yield good benchmark results that are driven by this property. Thus far, there is no evidence that real-world causal structures lead to varsortable data, and hence, our simulator of choice should reflect this fact. 
    \item \textbf{Distributional Properties}\label{des4}: Agreement between the properties of empirical and synthetic distributions is crucial to ensuring that novel algorithms perform as well in the real world as they do in theory.
\end{enumerate} These desiderata guide the design decisions described in Section \ref{sec:framework} where a different set of desiderata may have led to a different model.

\subsection{CausalRegNet} \label{sec:framework}

We propose a multiplicative effect SCM that accurately reflects the interaction dynamics of single-cell gene expression. Once parameterised, samples can be drawn efficiently from this simulator.

\subsubsection{Node-wise Distribution}

For each node $j$, the random variable associated with this node is distributed according to, \begin{equation}
    X_j \sim \text{NegBin}(\mu_j, \sigma_j^2),
\end{equation} where \(\mu_j\) is the mean expression and \(\sigma^2_j\) is the variance. The parameters are defined as \begin{equation}
    \mu_j = \mu_j^0 \cdot f_j^\text{reg}(X_{\text{pa}(j)}; \Theta_j), \;\;\;\; \text{ and } \;\;\;\;
    \sigma_j^2 = \mu_j(1 + \mu_j / \theta_j), 
\end{equation} where $\mu_j^0$ is the observational mean and $f_j^\text{reg}(\cdot)$ represents the regulatory effect of the parents of $X_j$ on $X_j$. A mean-variance relationship is imposed via the variance, $\sigma_j^2$, which is determined by the mean, $\mu_j$, and the inverse dispersion parameter, $\theta_j$. This functional form is based on well-established domain knowledge motivated by generic and well-known properties of biochemical interplay \citep{risso2018general}.

\subsubsection{Regulatory Function} \label{sec:regfunc}

The regulatory effect function takes a sigmoidal form, \begin{equation}
    f_j^\text{reg}(X_{\text{pa}(j)}; \Theta_j) = \frac{\alpha_j}{1 + \exp{\left\{-\gamma_j (w(x_{\text{pa}(j)}) + b_j)\right\}}},
\end{equation} where \(\alpha_j\) is the maximum regulatory effect of \(X_{\text{pa}(j)}\) on \(X_i\), and \(w(x_{\text{pa}(j)})\) is any function that aggregates the regulatory effect of parents. This regulatory function introduces non-linearity to the relationship between each node and its parents.

For simplicity, we use \(w(x_{\text{pa}(j)}) = \sum_{i \in \text{pa}(j)} w_{ij} \cdot x_i / \mu_i^0\), where \(W \in \mathbb{R}^{n\times n}\) is a weighted adjacency matrix that defines the strength of relationship between \(X_i\) and \(X_j\) and \([W]_{ij} = w_{ij}\) is the edge weight for the edge going from \(X_i\) to \(X_j\).

\subsubsection{Model Specification} \label{sec:modelspec}

The parameters \(\gamma_j\) and \(b_j\) are calibrated according to the aggregation function \(w(\cdot)\), maximum regulatory effect \(\alpha_j > 1\), and a minimum regulatory effect \(0 < \beta_j < 1\) parameter. To do this, we define the following value conditions: \begin{enumerate}[label=\Roman*]
    \item \textbf{Baseline Expression Condition (BEC):} \begin{equation*}
    \text{if } \forall i \in \text{pa}(j), X_i = 0, \text{ then } f_j^\text{reg}(X_{\text{pa}(j)}) = \beta_j,
    \end{equation*}
    \item \textbf{Observational Expression Condition (OEC):} \begin{equation*}
    \text{if } \forall i \in \text{pa}(j), X_i = \mu_i^0 \text{ then } f_j^\text{reg}(X_{\text{pa}(j)}) = 1.
    \end{equation*} 
\end{enumerate}


In other words, we first have that when the expression of each of the parents of node \(j\) are at zero, the regulatory effect will be at the baseline \(\beta_j\). Conversely, when each of the parents of \(j\) are expressed at their observational mean, then the regulatory effect is equal to 1, implying \(\mu_j = \mu_j^0 \cdot 1 = \mu_j^0\). That is, \(X_j\) is at its observational mean. 

These value conditions allow us to freely specify the aggregation function \(w(\cdot)\) and calibrate the regulatory effect function to guarantee maximum and minimum effects. Furthermore, these conditions are sufficient to yield closed-form solutions for \(\gamma_j\) and \(b_j\) given the parameters \(\{\alpha_j, \beta_j, w(\cdot)\}\).

Thus, given that the node-wise distribution, regulatory effect function and model specification method have clear biological interpretations, CausalRegNet satisfies \hyperref[des1]{Desiderata I}. 

\subsubsection{Example: Model Specification for Linear Aggregation with Mean-normalisation}

\begin{figure}
    \centering
    \includegraphics[width=0.5\textwidth]{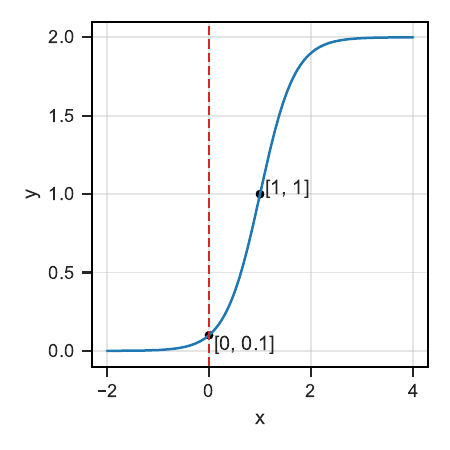}
    \caption{Example calibrated regulatory effect function. The red line indicates the cut-off, the regulatory effect will never go left of this line.}
    \label{fig:sigmoid_calibrated}
\end{figure}

Assuming the linear aggregation function introduced in Section \ref{sec:regfunc}, we demonstrate how value conditions can be used to derive the parameters \(\gamma_j\) and \(b_j\). Applying OEC and BEC from Section \ref{sec:modelspec}, we have that \(w(x_{\text{pa}(j)}) = 0\) when \(x_i = 0\), \(\forall i \in \text{pa(j)}\) and that \(w(x_{\text{pa}(j)}) = \sum_{i \in \text{pa}(j)} w_{ij}\) when \(x_i = \mu_i^0\), \(\forall i \in \text{pa}(j)\), respectively. Solving, this yields, \begin{equation}
    b_j = - \frac{1}{\gamma_j} \left( \log{\left( \alpha_j -1 \right)} + \gamma_j w^\prime_j \right) \;\;\;\; \text{and,} 
\end{equation} \begin{equation}
    \gamma_j = \frac{1}{w^\prime_j} \left[ \log{\left( \frac{\alpha_j}{\beta_j} -1 \right)} - \log{\left( \alpha_j -1 \right)} \right],
\end{equation} where \(w^\prime_j = \sum_{i\in\text{pa}(j)} w_{ij}\) and \(w^\prime_j \neq 0\). Assuming that \( w^\prime_j=1\) and setting the remaining parameters as \(\{\alpha_j = 2, \beta_j = 0.1\}\) the calibrated regulatory effect function is shown in Figure \ref{fig:sigmoid_calibrated}. The points in this figure highlight the OEC and BEC being satisfied on the regulatory effect curve.

\subsubsection{Adjusted Regulatory Function}

 When simulating nodes with excessively small observational means $\mu^0_j$, standard calibration as shown in Figure \ref{fig:sigmoid_calibrated} can lead to overly large aggregated regulatory effects. For example, suppose that we want to simulate a child of gene MTOR from \cite{replogle2022mapping} with $\hat{\mu}^0_j \approx 0.25$ and $\hat{\theta}_j \approx 3.38$ with $f^\text{reg}$ shown in Figure \ref{fig:sigmoid_calibrated}. The distribution of the expression of MTOR is shown in Figure \ref{fig:empirical-assessment-comb}(a), where it varies between 0 and 4. With such expression rates, we have that $x_i= 0, 1, 2, 3, 4$, which corresponds to $x_i / \mu_i^0 = 0, 4, 8, 12, 16$, respectively. This regulatory effect is much larger than the range of values for which the function in Figure \ref{fig:sigmoid_calibrated} varies. That is, for any $x_i > 1$, $f^\text{reg}_j(X_{\text{pa}(j)}) \approx 2$. Thus, we propose an adjusted regulatory function with an additional \textit{regulatory adjustment constant} $a_j$, \begin{equation}
    f^\text{reg}_j(X_{\text{pa}(j)}; \Theta_j) = \frac{\alpha_j}{1 + \exp{\{-\gamma_j (w_{\text{adj}}(x_{\text{pa}(j)}) + b_{\text{adj}})\}}}
\end{equation} where \begin{equation}
    b_{\text{adj}} = b_{j} - a_j, \;\;\;\; \text{and} \;\;\;\;  w_\text{adj}(x_{\text{pa}(j)}) = w(x_{\text{pa}(j)})+ a_j.
\end{equation} This additional hyperparameter now allows for full expression of regulatory effects even when $\mu_i^0$ is very small. This adjusted regulatory effect function is used for simulations throughout this work.

\subsubsection{Node-wise Distribution Fitting} \label{sec:modelfit}

By parametrising the distribution of each node as a Negative Binomial, we enable node-wise fitting of simulated distributions to empirical marginal distributions in real-world data. We fit the observational mean and inverse dispersion via the Negative Binomial likelihood \citep{risso2018general}. The implemented fitting scheme uses the L-BFGS-B to find a locally optimal solution for each node independently. Further details can be seen in Appendix \ref{sec:negbin}.

\section{Empirical Results} \label{sec:motivation}

The development of CausalRegnet was motivated by the lack of agreement between  interventional distributions generated by SERGIO and domain-knowledge in biology. We explore this motivation in Section \ref{sec:motivation1} alongside results from CausalRegNet. Guided by our desiderata, we then compare the scalability and varsortability of CausalRegNet, SERGIO and the Gaussian ANM in Sections~\ref{sec:scalability}-\ref{sec:varsortability}, respectively. Finally, in Section \ref{sec:comparison} we examine the agreement between CausalRegNet and real-world perturb-seq data. The full parameter settings used for simulation in this section are given in Appendix \ref{sec:sergioparameters}.

\subsection{Motivation: Interventional Distributions under Changing Structures} \label{sec:motivation1}

To understand simulator behaviour under intervention, we estimate the average treatment effects (ATEs) of the intervention $I=\{X_0\}$, on $X_1$, denoted \begin{equation} ATE(X_0, X_1) = \mathbb{E}\left[X_1^\emptyset\right] - \mathbb{E}\left[X_1^{\textit{do}(X_0 = 0)}\right],\end{equation} with varying causal structures, where $\emptyset$ denotes the observational setting.
    
\begin{figure}
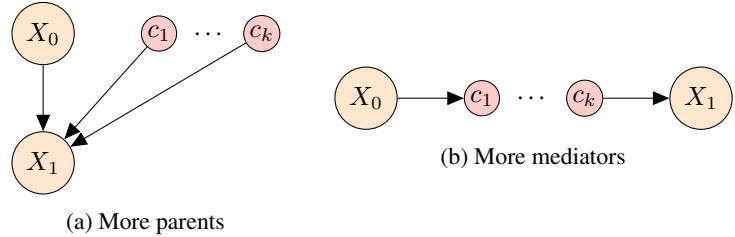

    \captionsetup[subfigure]{justification=centering}
\tikzstyle{latent} = [circle,fill=red!20,draw=black, inner sep=1pt, minimum size=10pt]
\tikzstyle{det} = [circle,fill=orange!20,draw=black, inner sep=3pt, minimum size=10pt]
    \centering
    \begin{subfigure}[c]{0.3\textwidth}
        \centering  
\tikz{ %
  \node[det] (x0) {${X_0}$} ; %
    \node[latent, right=\ltkiz of x0](c1){$c_1$};
      \node[det, below=\ltkiz of x0](x1){$X_1$} ;
    \node[latent, right=\ltkiz of c1](ck){$c_k$};
  
    \edge {x0} {x1};
    \edge {c1} {x1};
    \edge {ck} {x1};

    \path (c1) -- node[auto=false]{\ldots} (ck);

  }
  \caption{More parents}
  \label{fig:competing_study1}
    \end{subfigure}
    ~
    \begin{subfigure}[c]{0.4\textwidth}
        \centering   
\tikz{ %
  \node[det] (x0) {${X_0}$} ; %
    \node[latent, right=\ltkiz of x0](c1){$c_1$};
    \node[latent, right=\ltkiz of c1](ck){$c_k$};
      \node[det, right=\ltkiz of ck](x1){$X_1$} ;

    \edge {x0} {c1};
    \edge {ck} {x1};

    \path (c1) -- node[auto=false]{\ldots} (ck);
  }         
  \caption{More mediators}
  \label{fig:competing_study2}
    \end{subfigure}
    \caption{We examine the treatment effect of $X_0$ on $X_1$ as (a) the number of parents of $X_1$ increases and (b) as the number of mediators between $X_0$ and $X_1$ increases (i.e. for $k=0,\ldots, 9$ under (a) and (b)).}
\label{fig:competing_study} 
\end{figure}

First, we study the ATE as the number of parents of a node increases. That is, we are interested in how the distribution of a random variable changes as the number of parents increases. Figure~\ref{fig:competing_study1} illustrates this setting, where the number of parents, $k$, varies. 

\begin{figure}[ht]
    \centering
    \includegraphics[width=0.9\textwidth]{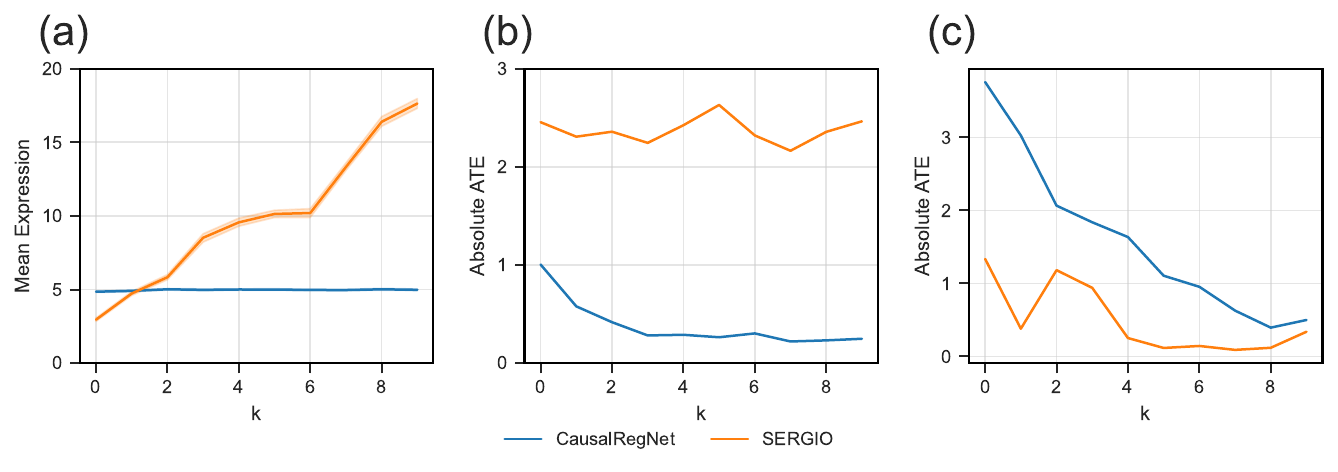}
    \caption{(a) shows absolute ATE with increasing numbers of parents. In (b), the mean observational expression of \(X_1\) against number of parents. (c) shows the absolute ATE falls for both simulators as the number of mediators increases.}
    \label{fig:competing_study_res}
\end{figure}

In Figure \hyperref[fig:competing_study_res]{3a}, simulating with SERGIO gives an average ATE of 2.5 across simulations, whilst the mean expression of $X_1$ is rising. This suggests that the constant ATE is facilitated by the increasing expression level, allowing each additional parent to have an ATE of 2.5. Conversely, CausalRegNet with the mean-normalised aggregation function applied in Section \ref{sec:modelspec} ensures that each regulator has equal contribution, and thus, as the number of parents increases, the observed ATE tends towards zero. Although it is not clear what aggregation function is most faithful to the underlying biology, the modularity of CausalRegNet means that the aggregation function can be adapted to emerging domain knowledge. For both CausalRegNet and SERGIO, the ATE is partially determined by the weighted adjacency matrix, such that parents with greater edge weights will determine more of the regulatory effect and will result in different ATEs. This can be directly altered when using either simulator. 

Next, we consider the case where the number of nodes mediating the effect of $X_0$ on $X_1$ increases. Figure \ref{fig:competing_study2} illustrates this setting, where $k=0,\ldots,9$. Figure \hyperref[fig:competing_study_res]{3c} shows that as the number of mediators increases, the ATE falls. This is intuitively expected since ancestors further away may plausibly lead
to smaller changes in downstream nodes. Thus, both simulators behave as expected under this case.

\subsection{Scalability} \label{sec:scalability}

\begin{figure}[h!]
    \centering
    \includegraphics[width=0.8\textwidth]{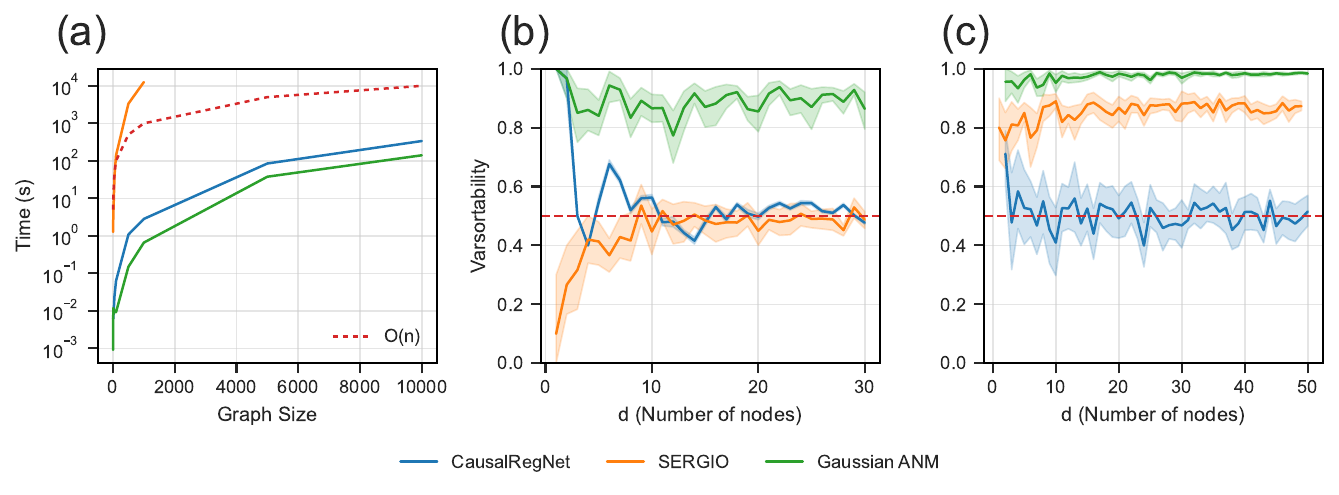}
    \caption{(a) Simulation time for data from graphs of increasing size. Data is generated from the same causal graph structures across methods. Computational time for a single simulation with SERGIO with 5,000 nodes or more exceeds 24 hours and so is not included. Mean varsortability of data generated from (b) a causal chain and (c) a causal graph structure by each simulator with 95\% confidence intervals.}
    \label{fig:full_bench}
\end{figure}

Figure \hyperref[fig:full_bench]{4a} shows the time taken to simulate data from predetermined graphs of size 3 to 10,000. The scalability of CausalRegNet is shown to be comparable to the simple Gaussian ANM, which can generate data from graphs of size 1,000 within less than 10 seconds. For the same graphs, SERGIO takes upwards of 10,000 seconds or 2.78 hours. Moreover, SERGIO takes over 100,000 seconds or 24 hours to simulate a causal graph of 5,000 or more nodes, which makes it practically 
very difficult to simulate such large datasets. 

The poor scalability of SERGIO is caused by the need to reach a steady state before samples can be drawn. Methods based on SCMs such as CausalRegNet or any ANM are defined by parametric probability distributions and can thus be be drawn from instantaneously. The simulation time in this instance is purely associated with the time needed to traverse the causal graph according to the topological ordering of nodes. 

Thus, CausalRegNet satisfies \hyperref[des2]{Desiderata II} as it is scalable to the dimension of real-world experiments.

\subsection{Varsortability} \label{sec:varsortability}

To examine the varsortability of data generated by each simulator we generated data from causal chain and causal graph structures. A causal chain is a causal graph such that only one directed path of the form \(X_{1} \rightarrow \ldots \rightarrow X_{n}\) exists, starting from the first node in the chain, \(X_{1}\), to the last \(X_{n}\). For CausalRegNet, we use nodes fitted to real gene expression data from \cite{replogle2022mapping} by randomly selecting genes from the set obtained in Appendix \ref{sec:exampledist}. Parametric settings for simulating from SERGIO can be found in Appendix \ref{sec:sergioparameters}. Parameters used to generate causal graphs used by both CausalRegNet and SERGIO are in Appendix \ref{sec:daggeneration}.

The results for the causal chain simulations are shown in Figure \hyperref[fig:full_bench]{4b}. For causal chains of length 10 or more, CausalRegNet and SERGIO generate near perfectly non-varsortable data. In Figure \hyperref[fig:full_bench]{4c} CausalRegNet generates data with very low varsortability and outperforms both SERGIO and the Gaussian-ANM in this respect. Further, Table \ref{tab:varsort} shows the average varsortability of data generated by the causal chains and graphs with the greatest number of nodes (i.e. $d=\{30, 50\}$, respectively), with CausalRegNet achieving significantly lower varsortability than both SERGIO and the Gaussian ANM. 

CausalRegNet therefore satisfies \hyperref[des3]{Desiderata III} as it generates data with low varsortability.

\begin{table}
\centering
\caption{Varsortability of data generated by the largest causal chains ($d=30$) and causal graphs ($d=50$) with 95\% confidence intervals.}
\label{tab:varsort}
\begin{tabular}{ccc}
\hline
Simulator                     & Causal Chain Varsort. & Causal Graph Varsort. \\ \hline
Gaussian ANM                  & $0.863 \pm 0.076$           & $0.986 \pm 0.005$           \\
SERGIO                        & $\mathbf{0.488 \pm 0.034}$           & $0.873 \pm 0.019$           \\
CausalRegNet  & $\mathbf{0.477 \pm 0.008}$           & $\mathbf{0.492 \pm 0.008}$    \\  
\hline
\end{tabular}%
\end{table}

\subsection{Comparison to Experimental Data} \label{sec:comparison}

To conduct an empirical comparison between the data generated by CausalRegNet and real-world data, we collect 100 genes related to cancer pathways (see Appendix \ref{sec:exampledist}) for the most commonly referenced genes that were both measured and perturbed in the experimental dataset from \cite{replogle2022mapping}.

Given this set of 100 genes, we aim to simulate graphs of size $d=\{3, 10, 100\}$ with CausalRegNet to assess whether the simulated distributions are similar to those observed in \cite{replogle2022mapping}. Using the distribution fitting scheme outlined in Section \ref{sec:modelfit}, each node in the DAG is fitted to simulate the expression of a gene from the original set of 100 genes.

\begin{figure}
    \centering
    \includegraphics[width=0.78\textwidth]{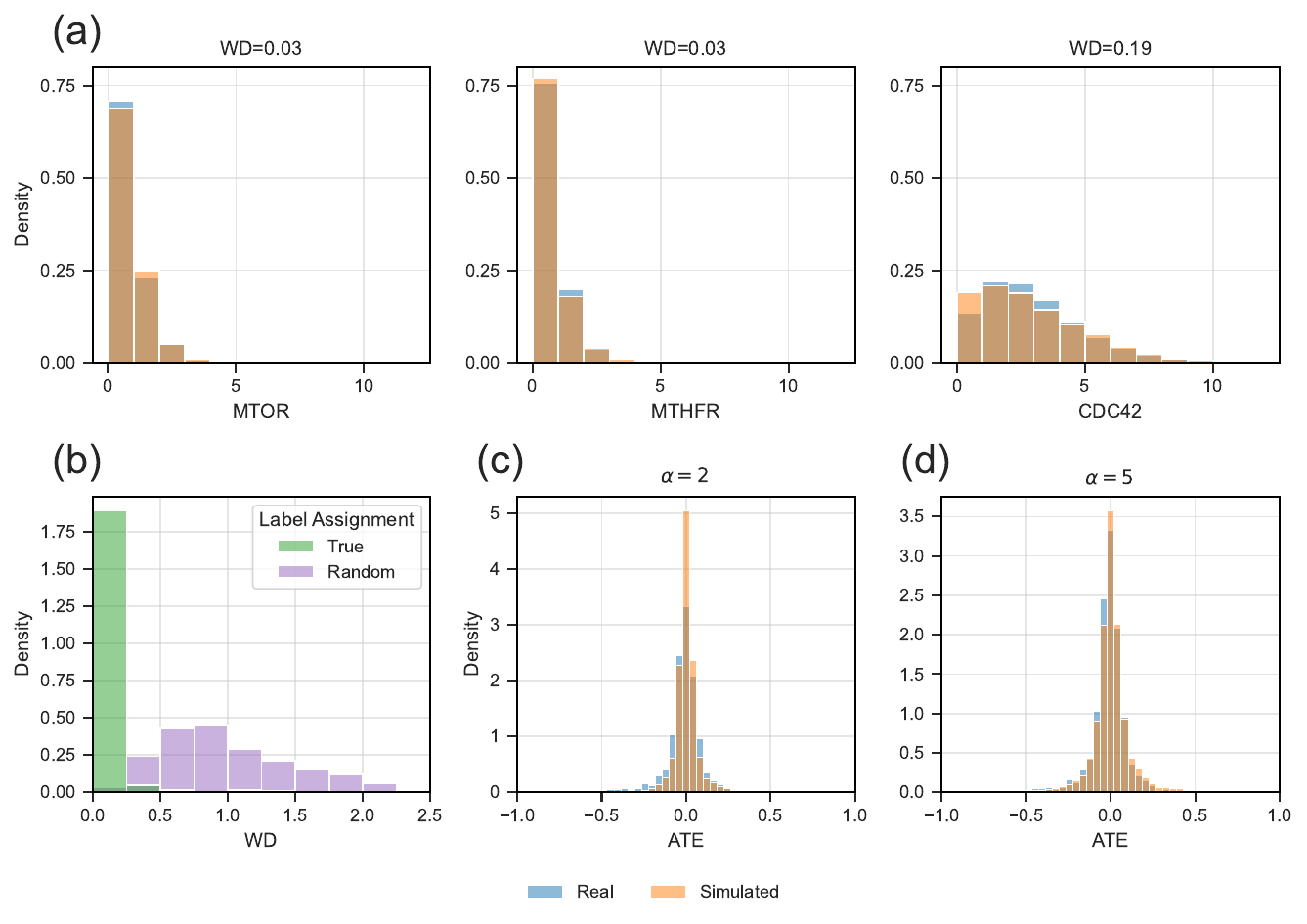}
    \caption{(a) Comparison between real and simulated data for nodes fitted to genes in \cite{replogle2022mapping} in DAG with 3 nodes. (b) In green, WD for simulated marginal distributions fitted to \cite{replogle2022mapping} compared to the true distributions. In purple, the distribution labels are shuffled such that each simulated marginal is compared to a random empirical marginal, giving a baseline for comparison. The distribution of interventional effects with (c) $\alpha_j = 2$ and (d) $\alpha_j=5$ for each node.}
    \label{fig:empirical-assessment-comb}
\end{figure}

Figure \hyperref[fig:empirical-assessment-comb]{5a} shows the real and simulated nodes for 3 genes selected from \cite{replogle2022mapping}. We measure the distance between simulated and real distributions using the Wasserstein distance (WD) \citep{ramdas2017wasserstein, 2020SciPy-NMeth}. The overlap between simulated and real distributions is $ < 0.2$ for all nodes. Figure \hyperref[fig:empirical-assessment-comb]{5b} shows the WDs for graphs of size $d=100$. As a comparative baseline, we shuffle the gene labels such that each synthetic distribution is compared to the expression of a random gene. From Figure \hyperref[fig:empirical-assessment-comb]{5b}, clearly the comparison to the true distribution leads to lower WDs. For both true and random label assignments, there are also outliers in the WDs, shown in Figure \ref{fig:emp_violin} (Appendix \ref{sec:wdapp}), however, these account for a small number of nodes. To assess the synthetic interventional distributions, since the true causal graph is unknown we cannot compare distributions directly. Instead, in Figures \hyperref[fig:empirical-assessment-comb]{5c-d} we compare the \textit{distribution} of interventional effects between the real and simulated data. We can see that we obtain good coverage with some excessive $0$ effects in \hyperref[fig:empirical-assessment-comb]{5c}, however, modifying $\alpha_j$ leads to a smaller number of $0$ as shown in Figure \hyperref[fig:empirical-assessment-comb]{5d}. 

Hence, CausalRegNet also satisfies \hyperref[des4]{Desiderata IV} as it generates observational and interventional distributions that are comparable to real-world data.

\subsection{Benchmark Study} \label{sec:benchmarkmain}

The goal of this work is to introduce a context-specific simlulator for use in studying CSL. To demonstrate its use,
we carried out an \textit{illustrative} study benchmarking some CSL algorithms commonly used as baselines (see Appendix \ref{sec:benchmark}). Here, we see that in the specific context used, the various methods perform poorly. We emphasise that fully leveraging our simulators for benchmarking {\it per se} goes beyond the scope of this paper and would be a key direction for future work.

\section{Discussion} \label{sec:discussion}

In this work, we proposed CausalRegNet, a simple yet flexible SCM-based approach to simulation in the context of large-scale gene perturbation screens. We found that CausalRegNet led to significantly improved scalability over existing
approaches whilst coping with the issue of varsortability and 
providing synthetic data matched to key properties of real data distributions. Recent work \citep{ng2024structure} has shown that varsortability corresponds to a necessary assumption for some ANMs, as shown by \cite{loh2014high}, thus CausalRegNet again provides an approach for comparing methods with varying assumptions under a fair, realistic setting.
In addition to its use for benchmarking and context-specific assessment, 
we think CausalRegNet 
can be a useful tool in 
training machine learning approaches for causal discovery and inference by allowing for realistic and rapid simulation. Rapid simulation frameworks can also contribute to the design of real-world experiments, including active learning schemes \cite{sussex2021near, tigas2022interventions, tigas2023differentiable}.

 Individual existing simulators satisfy some but not all of the desiderata set out in this work. SERGIO is able to generate accurate observational distributions but lacks the ability to realistically simulate interventions and scalability to 
 large  problems. GRouNdGAN generates accurate observational distributions and is scalable but lacks interpretability and has not yet been assessed on its ability to simulate interventions. GRouNdGAN also requires training and validation on a reference dataset to accurately simulate data, which can be computationally expensive. In contrast, we provide a library of parameters derived from \citep{replogle2022mapping} that allows simulations to be carried out parsimoniously.

\paragraph{Limitations \& Future Work.} Several properties of the framework and tool here require further work to facilitate development of effective context-specific benchmarking tools. The simplicity of the SCM-based approach means that temporal dynamics cannot be simulated, which is possible in less scalable tools such as SERGIO and approaches that allow time-dependent dynamics simulation 
will be important in some applications of CSL. Further work is also needed to explore alternative parameterisations of the aggregation functions \(w(\cdot)\). A key direction for future empirical work will be to conduct a comprehensive benchmarking study using CausalRegNet to 
compare the performance of a library of existing CSL algorithms.

\begin{ack}
LK acknowledges the support of an Ivan D Jankovic Studentship at Clare Hall, University of Cambridge. This work was funded by UKRI Programme Grant MC\_UU\_00002/18. For the purpose of open access, the author has applied a Creative Commons Attribution (CC BY) licence to any Author Accepted Manuscript version arising.

\color{blue}

\end{ack}

\bibliographystyle{unsrt}
{\footnotesize
\bibliography{references}}

\appendix

\part*{Appendices}

In Appendix \ref{sec:scrnaseq}, we introduce single-cell RNA sequencing experiments and the biology necessary for CausalRegNet. Appendix \ref{sec:sergio} includes details on SERGIO further to those included in the main paper. The likelihood used to fit the negative binomial distributions of CausalRegNet to real data is given in Appendix \ref{sec:negbin}. Finally, Appendix \ref{sec:further_results} includes a benchmark on commonly used CSL algorithms with some additional empirical evaluations of CausalRegNet.

\section{Introduction to Single-cell RNA Sequencing} \label{sec:scrnaseq}

Genes are DNA sequences that are transcribed into messenger RNA (mRNA) before being turned into proteins. The level of mRNA corresponding to each gene within a particular cell is referred to as the \textit{gene expression} and is a measure of the activity of that gene in that specific context. Gene expression is highly plastic, varying according to the type of cell, tissue, time and internal and external stimuli, including the expression of other genes. Many disease associated DNA variants lie outside genes but seem to exert their effects by influencing RNA expression at genes. Gene expression can be experimentally measured within individual cells, in \textit{single-cell} RNA sequencing (scRNA-seq) experiments, or aggregated across cells within a tissue sample, in \textit{bulk} RNA-seq experiments. This in turn allows for the study of the regulatory structure governing gene expression, enabling us to identify and analyse gene-gene interactions. 

Experimental techniques for collecting scRNA-seq data introduce additional \textit{technical noise}, which obscures the already noisy biological signal present in scRNA-seq data. Existing statistical approaches for studying scRNA-seq data often assume that data follows a (zero-inflated) Negative Binomial distribution and have been shown to be effective for signal extraction and distilling biological and technical noise \citep{risso2018general}. 

\section{Further details on SERGIO} \label{sec:sergio}

SERGIO (single-cell expression of genes \textit{in-silico}) is a technique for simulating scRNA-seq dynamics using stochastic differential equations (SDEs). Given a GRN structure, SERGIO simulates gene expression dynamics as a function of its regulators. It does so using the chemical Langevin equation (CLE) (cite) where the time-course mRNA concentration of gene $i$ is given by: \begin{equation}
    \frac{dx_i}{dt} = P_i(t) - \lambda_i x_i(t) + q_i \left( \alpha \sqrt{P_i(t)} + \beta \sqrt{\lambda_i x_i(t)} \right)
\end{equation} where $x_i(t)$ is the expression of gene $i$, and $P_i(t)$ is the production rate at time $t$. Noise is added through the independent Gaussian white noise processes $\alpha$ and $\beta$. The production rate is modelled as a sum of contributions from each parent: \begin{equation}
    P_i = \sum\limits_{j\in\text{pa}(i)} p_{ij} + b_i
\end{equation} where $p_{ij}$ is the regulatory effect of parent $j$ on gene $i$ and $b_i$ is the basal production rate of gene $i$. $p_{ij}$ is determined by a non-linear function of the expression of the parent according to: \begin{equation}
    p_{ij} = K_{ij} \frac{x_j^{n_{ij}}}{h_{ij}^{n_{ij}} + x_j^{n_{ij}}}
\end{equation}
if parent $j$ is an activator and  \begin{equation}
    p_{ij} = K_{ij}\left( 1 - \frac{x_j^{n_{ij}}}{h_{ij}^{n_{ij}} +x_j^{n_{ij}}} \right)
\end{equation} if parent $j$ is a repressor. Here, $K_{ij}$ denotes the maximum contribution of regulator $j$ to target $i$,  $n_{ij}$ is the Hill coefficient that modulates non-linearity, and $h_{ij}$ is the half-maximal effect.  

 The expression of gene $i$ is updated according to: \begin{equation}
     (x_i)_{t+\Delta t} = (x_i)_t + (P_i(t) - \lambda_i x_i(t)) \Delta t + q_i \sqrt{P_i(t)} \Delta W_\alpha + q_i \sqrt{\lambda_i x_i(t)} \Delta W_\beta
 \end{equation} where $\Delta W_\alpha \sim \sqrt{\Delta t} N(0,1), \Delta W_\beta \sim \sqrt{\Delta t} N(0, 1)$. See \cite{dibaeinia2020sergio} for full details. 

Here, we modify SERGIO to perform hard interventions by setting \(x_i = a\) for the $i$-th node, as done by other applications of SERGIO \cite{hagele2023bacadi}.

\section{Negative Binomial Likelihood} \label{sec:negbin}

As conducted by \cite{risso2018general}, the Negative-Binomial likelihood used to fit the observational mean \(\mu_j^0\) and inverse dispersion \(\theta_j\) for node \(j\) is given by,

\begin{equation}
    f_\text{NB}(x_j;\mu^0_j, \theta_j) = \frac{\Gamma(x_j+\theta_j)}{\Gamma(x_j + 1) \Gamma(\theta_j)} \left(\frac{\theta_j}{\theta_j + x_j}\right)^{\theta_j}\left(\frac{\mu_j}{\mu_j + \theta_j}\right)^{x_j}. 
\end{equation}

The log-likelihood \(\log{f_\text{NB}(x_j; \mu_j^0, \theta_j)}\) is optimised using the L-BFGS-B implementation in SciPy \citep{2020SciPy-NMeth}.

\section{Extended Empirical Results} \label{sec:further_results}

In this section, we present initial benchmarking results of commonly used CSL methods using CausalRegNet and a description of the computational resources needed to perform these experiments in Section \ref{sec:compute-resources}. We start by describing the parameter settings used for simulations involving CausalRegnet and SERGIO in Section \ref{sec:sergioparameters}. Section \ref{sec:benchmark} then shows full simulation results. The methodology used for selecting the 100 cancer-related genes for CausalRegNet is given in Section \ref{sec:exampledist}. A comparison of the distribution of correlations in the CausalRegNet synthetic dataset and Replogle as well as for the causal and non-causal relationships simulated by CausalRegNet are shown in Section \ref{sec:corr}. The distribution of Wasserstein distances between synthetic and real gene distributions for the true and random gene labels is given in Section \ref{sec:wdapp}. 

\subsection{Computational Resources} \label{sec:compute-resources}

CPU hours were used to run all simulations. The most intense experiments involved generating data using SERGIO for the scalability results in Section \ref{sec:scalability}. This involved over 300 compute hours with over 64GB of active memory. The remaining experiments in Section \ref{sec:varsortability} and Section \ref{sec:comparison} were run locally with basic compute requirements. The experiments involving CSL algorithms also exceeded over 150 compute hours to run the algorithms on graphs of size 50 and 100 with 16GB of active memory.

\subsection{Simulation Parameter Settings} \label{sec:sergioparameters}

For CausalRegNet, individual node means and inverse dispersions are obtained from fitting a negative binomial distribution to randomly selected genes in Section \ref{sec:exampledist}. The minimum effect, maximum effect and regulatory adjustment effect for each node are $\{\alpha_j = 2, \beta_j=0.1, a_j=10\}$, respectively, and edge weights are sampled from the range $w_{ij}\in (-2, -0.5) \cup (0.5, 2)$. For SERGIO, we follow the default parameters specified by \cite{dibaeinia2020sergio}, which are given in Table \ref{tab:sergio-sim-params}. The MR is the master regulator of each causal graph, that is, the node with no parents in each graph. DAGs are randomly generated according to Table \ref{tab:sim-params} and identical DAGs are used for both CausalRegNet and SERGIO. \begin{table}[h!]
\centering
\caption{Parameters for synthetic data generation with SERGIO. }
\label{tab:sergio-sim-params}
\begin{tabular}{@{}l|l@{}}
\toprule
Interaction Strength                     & \(k_{ij} \sim \text{Unif}(1,5)\) \\
MR Basal Production                       & \(b_i \sim \text{Unif}(5, 15)\)               \\
Noise Amplitude & \(q = 1\)              \\
Decay       & \(\lambda=0.8\)                                  \\
Hill Coefficient (degree of non-linearity)        & \(n_{ij} = 2\)                                 \\ \bottomrule 
\end{tabular}%
\end{table}

\begin{table}[h!]
\centering
\caption{Parameters for DAG generation.}
\label{tab:sim-params}
\begin{tabular}{@{}l|l@{}}
\toprule
Repetitions                      & 10                                                \\
Graphs                           & ER-$m$                                           \\
Nodes                            & \(d \in \{10, 20\}\)              \\
Edges                            & \(m \in \{2 \cdot d, 4 \cdot d\}\)              \\
Samples Per Setting                        & \(n=1000\)                                        \\
\bottomrule 
\end{tabular}%
\end{table}

\subsection{Benchmark Study} \label{sec:benchmark}

Here, we conduct a simulation study to examine the performance of various CSL algorithms using CausalRegNet as our synthetic data generator. 

Our simulation study includes a diverse set of CSL methods. The PC algorithm, introduced by \cite{spirtes2000causation}, is a \textit{constraint-based} method, utilising conditional independence testing to identify the causal strucuture with observational data. \cite{zheng2018dags} developed a continuous optimisation scheme on the space of causal graphs through NO-TEARS. Finally, we include the GSP and IGSP methods based on traversing the space of sparsest permutations, which were developed by \cite{solus2021consistency} and \cite{wang2017permutation}, respectively. IGSP is an interventional method and thus will be provided both observational and interventional data. Each method is implemented according to the settings shown in Table \ref{tab:benchmark_param}. 

\subsubsection{Setting} \label{sec:daggeneration}

We use a normalised Structural Hamming Distance (SHD), which we define as \(\text{SHD} / (m \cdot d)\), false omission rate (FOR) and false discovery rate (FDR), to evaluate the performance of each method on full graph retrieval. Both the Erd\H{o}s-R\'{e}nyi (ER) and Scale-Free (SF) graph strucures methods are used. The full parameters for generating DAGs that are used to generate data with CausalRegNet are shown in \ref{tab:sim-params}. For observational methods, data coming from only the observational setting, denoted by \(\mathcal{I} = \emptyset\), is provided.

\begin{table}[h!]
\centering
\caption{Parameter settings used for CSL algorithms for the benchmark.}
\label{tab:benchmark_param}
\begin{tabular}{@{}cl@{}}
\toprule
\multirow{3}{*}{PC} & \verb|indep_test = kci| \\
 & \verb|stable| = \verb|True| \\
 & \verb|alpha = 0.05| \\ \midrule
\multirow{2}{*}{NO-TEARS} & \verb|class = mlp| \\
 & \verb|max_iter = 20| \\ \midrule
\multirow{3}{*}{GSP} & \verb|ci_test| = \verb|kci| \\
 & \verb|alpha = 1e-3| \\
 & \verb|depth = 5| \\ \midrule
\multirow{4}{*}{IGSP} & \verb|ci_test| = \verb|kci| \\
 & \verb|inv_test = kci| \\
 & \verb|alpha = 1e-3| \\
 & \verb|depth = 5|\\ \bottomrule
\end{tabular}
\end{table}

\subsubsection{Results}

\begin{figure}
\begin{subfigure}[h]{0.25\linewidth}
\includegraphics[width=\textwidth]{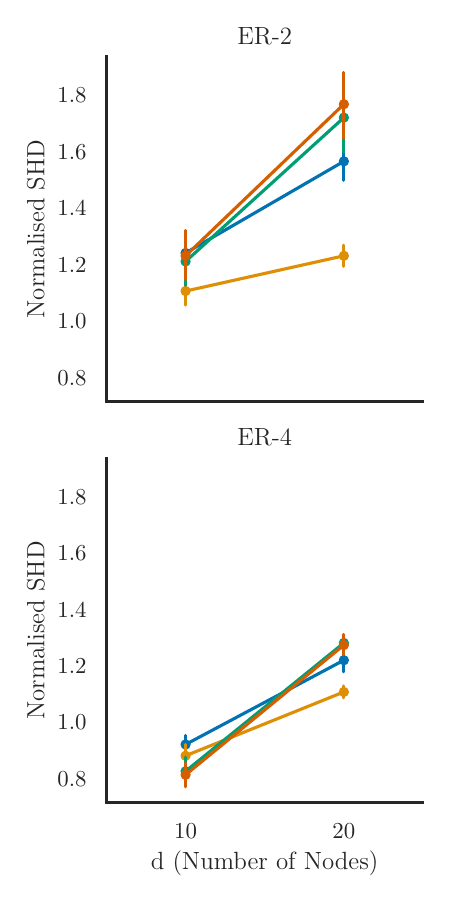}
        \caption{SHD performance for various causal graph settings.}
        \label{fig:shd}
\end{subfigure}%
\hfill
\begin{subfigure}[h]{0.25\linewidth}
    \includegraphics[width=\textwidth]{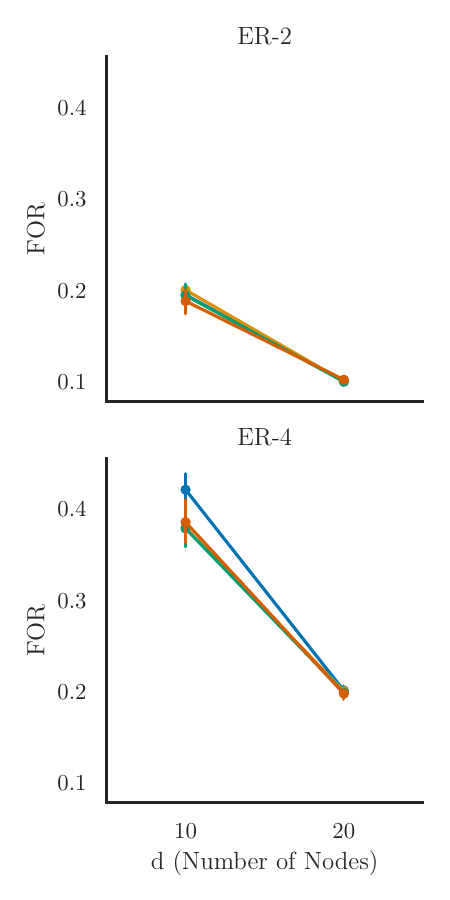}
    \caption{FOR performance for various graph settings.}
    \label{fig:for}
\end{subfigure}%
\hfill
\begin{subfigure}[h]{0.36\linewidth}
\includegraphics[width=\textwidth]{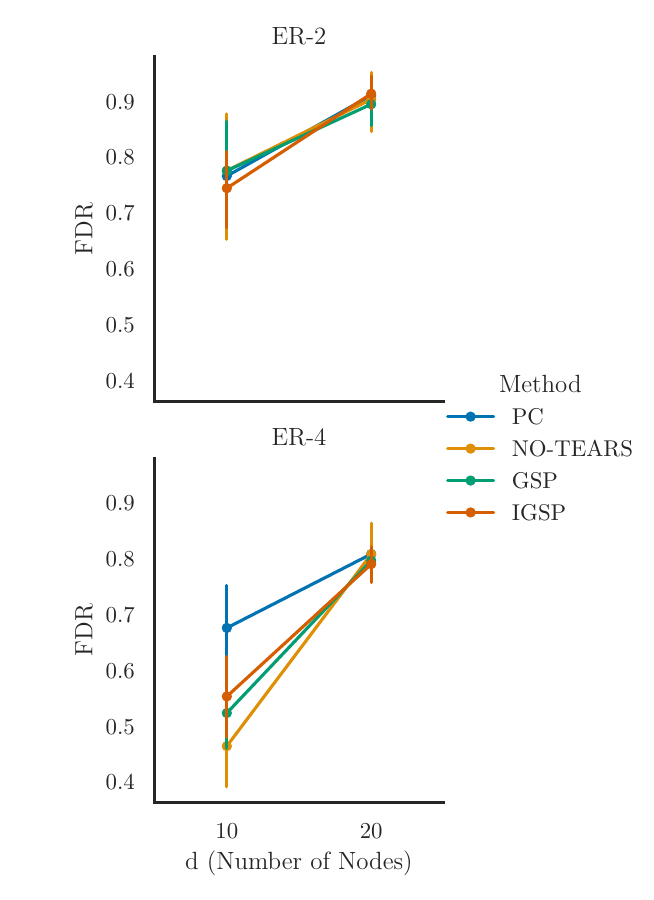}
        \caption{FDR performance for various graph settings.}
        \label{fig:fdr}
\end{subfigure}
\caption{Results from benchmark with CausalRegNet.}
\end{figure}

Figure \ref{fig:shd} shows that the normalised SHD for all methods increases with graph size, suggesting that the size of the graph significantly increases the difficulty of CSL.  Furthermore, this trend corresponds to decreasing specificity of each of the CSL algorithms as shown by increasing FDR with graph size in Figure \ref{fig:fdr}. Figure \ref{fig:for} shows that as the graph size grows, all methods are less likely to omit edges incorrectly, supporting our hypothesis of falling specificity. NO-TEARS consistently outperforms the other methods.

This benchmark highlights that despite the theoretical foundations of methods in CSL, there exist settings where provably correct assumptions, such as causal sufficiency, and causal faithfulness are not enough to ensure good performance. Thus, the need for testing the sensitivity of methods to violations of modelling assumptions in a context-specific manner is crucial.

\begingroup
\setlength{\tabcolsep}{15pt} 
\renewcommand{\arraystretch}{1.0} 

\endgroup

\subsection{Selecting Cancer Genes} \label{sec:exampledist}

We utilised  Entrez Direct tool \cite{kans2013entrez} and Pangaea \cite{pirvan2020pangaea} to parse abstracts from PubMed articles using the search term "cancer". From these abstracts, we identified the 100 most commonly referenced cancer-related genes, which were both perturbed and measured in the Replogle dataset. 

\subsection{Correlation Patterns} \label{sec:corr}

\begin{figure}
     \centering
     \begin{subfigure}[b]{0.45\textwidth}
         \centering
          \includegraphics[width=\linewidth]{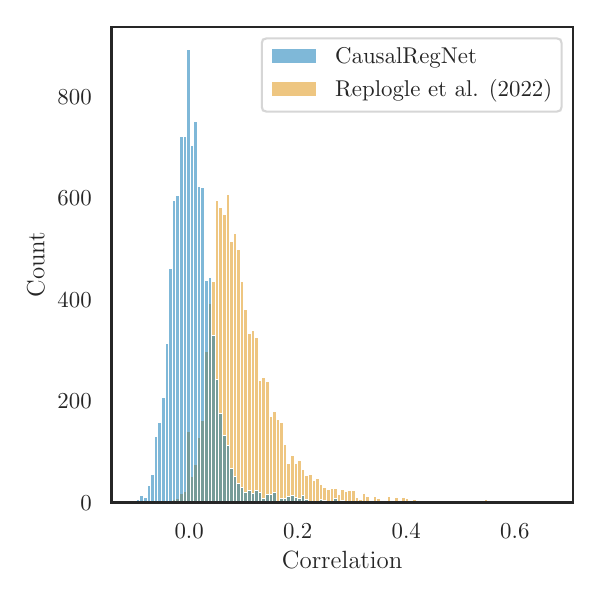}
          \caption{Distribution of correlations between genes in data simulated by CausalRegNet and real data from \cite{replogle2022mapping}.}
          \label{fig:corr_dist}
     \end{subfigure}
     \hfill
     \begin{subfigure}[b]{0.45\textwidth}
         \centering
         \includegraphics[width=\linewidth]{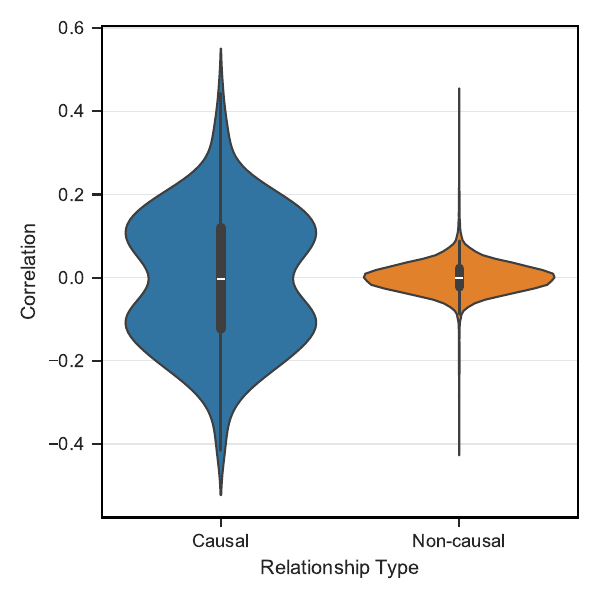}
          \caption{Distribution of correlations between nodes that have a direct causal correlation and those that do not.}
          \label{fig:causal_corr}
     \end{subfigure}
     \caption{Correlation patterns in data generated by CausalRegNet.}
\end{figure}

Furthermore, Figure \ref{fig:corr_dist} shows the correlation between 100 genes simulated by CausalRegNet and their real-world correlation in \cite{replogle2022mapping}. The overall distributions of correlation have similar shape where they are relatively centered about some mean correlation value. The real data shows a positive skew, which suggests that there is more upregulation than downregulation occurring in this particular set of genes than expected. For this simulation, CausalRegNet assumes a balanced distribution of upregulators ($w_{ij} > 0$) and downregulators ($w_{ij} < 0$). Some tweaking of this balance may be needed to achieve a more comparable distribution of correlations. This skew may also be due to all genes studied here being cancer-related. Therefore, they may be more correlated in expression than would be expected with a random subset of genes taken from \cite{replogle2022mapping}. Further work would be needed to verify this.

We also look at the difference in correlation between nodes that have a \textit{direct} causal relationship and those that do not. Figure \ref{fig:causal_corr} shows that there is a clear relationship between causal links and correlation.

\newpage

\subsection{Wasserstein Distance between Real and Synthetic Data} \label{sec:wdapp}

\begin{figure}[h!]
    \centering
    \includegraphics[width=0.4\textwidth]{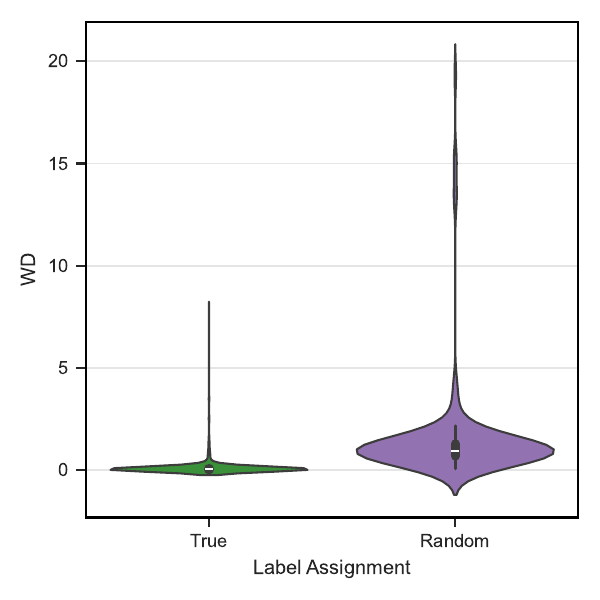}
    \caption{Distribution of Wasserstein distances }
    \label{fig:emp_violin}
\end{figure}

The results shown in Section \ref{sec:comparison} show that simulating data with fitted negative binomial parameters in CausalRegNet leads to better than random marginal distribution fit when compared to real data. Although this is true for the majority of nodes, there can still be outliers when simulating large graphs as shown in Figure \ref{fig:emp_violin}. These outliers occur at a constant rate thus for smaller graphs, a smaller number of poorly fitting nodes will appear. 

\end{document}